\documentclass[%
reprint,
superscriptaddress,
groupedaddress,
amsmath,amssymb,
floatfix
]{revtex4-2}

\newcommand{\pe}{\texttt{PlotExtract} }

\newcommand{\papertitle}{
Leveraging Vision Capabilities of Multimodal LLMs for Automated Data Extraction from Plots}

\usepackage{graphicx}
\usepackage{dcolumn}
\usepackage{xcolor}
\usepackage{colortbl}
\usepackage{booktabs}
\definecolor{ggrey}{rgb}{0.5,0.5,0.5}
\definecolor{gggrey}{rgb}{0.9,0.9,0.9}
\usepackage{multirow}
\usepackage{makecell}
\usepackage{bm} 
\usepackage{hyperref} 
\usepackage[mathlines]{lineno} 
\usepackage{float}
\usepackage{soul}
\usepackage{lipsum}
\usepackage{multirow}
\usepackage{bold-extra}

\usepackage{fancyhdr}

\begin{document}

\title{\papertitle}

\author{Maciej P. Polak}
\email{mppolak@wisc.edu}
\affiliation{Department of Materials Science and Engineering, University of Wisconsin-Madison, Madison, Wisconsin 53706-1595, USA}
\author{Dane Morgan}
\email{ddmorgan@wisc.edu}
\affiliation{Department of Materials Science and Engineering, University of Wisconsin-Madison, Madison, Wisconsin 53706-1595, USA}

\begin{abstract}

Automated data extraction from research texts has been steadily improving, with the emergence of large language models (LLMs) accelerating progress even further. Extracting data from plots in research papers, however, has been such a complex task that it has predominantly been confined to manual data extraction. We show that current multimodal large language models, with proper instructions and engineered workflows, are capable of accurately extracting data from plots. This capability is inherent to the pretrained models and can be achieved with a chain-of-thought sequence of zero-shot engineered  prompts we call \pe, without the need to fine-tune. We demonstrate \pe here and assess its performance on synthetic and published plots. We consider only plots with two axes in this analysis. For plots identified as extractable, \pe finds points with over 90\% precision (and around 90\% recall) and errors in \emph{x} and \emph{y} position of around 5\% or lower. These results prove that multimodal LLMs are a viable path for high-throughput data extraction for plots and in many circumstances can replace the current manual methods of data extraction.

\end{abstract}

\maketitle

\section{Introduction}
\begin{figure*}
    \centering
    \includegraphics[width=0.9\textwidth]{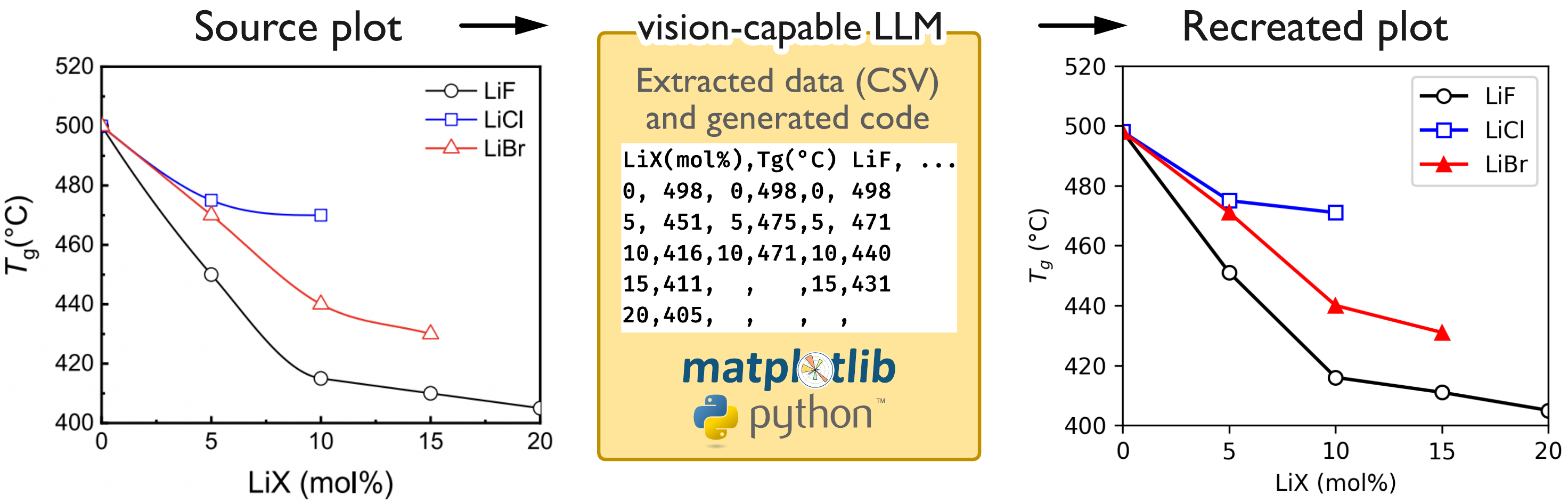}
    \caption{An example of plot data extraction. The source plot is Fig. 19 in \cite{example_fig19}}.
    \label{fig:example}
\end{figure*}

Plots in research documents contain a significant amount of data that does not exist in the document's text, and while data can now be efficiently extracted from text with the use of LLMs \cite{llm:chatextract,llm:extraction_review}, automated and unsupervised extraction of data from research plots remains an open challenge. Recent development in multimodal LLMs, and their vision capabilities in particular, allow for a significant progress in this area. 
In general, identification of a figure or a plot that contains information relevant to a topic is not very challenging. Most often, just the caption of the figure or a description of the figure in the main text of the paper provides sufficient information \cite{llm:chatextract}. Extracting the data from the figure, however, poses an entirely different challenge. For figures containing images, such as those from SEM or TEM microscopy \cite{llm:microscopy_vision}, or bio- and medical imaging \cite{llm:bio_vision,llm:bio_vision2,medical2,medical_review}, LLMs have been shown to be able to identify certain features or classifications. Similarly, figure identification and classifying its content using LLMs has been demonstrated in chemistry texts \cite{reticular}. Complex workflows, where text analysis is combined with figure identification and extracting certain features from the figures \cite{mermaid} have been recently developed as well.
On the other hand, for plots with points or curves representing data, data is typically still extracted manually \cite{manual_plots, manual2}. A typical procedure for such data extraction, assuming the plot is a raster image (which is the most common scenario) consists of pixel-wise identification of the location of the axes, and the ranges they span (more strictly, finding at least two identifying features on each axis with their corresponding  coordinates in the plot's space to determine their direction and establish the pixel size in the plot's space). Then, a pixel-wise identification of the data (points or curves) is manually performed by identifying corresponding pixels and correlating their coordinates to the plot's space coordinates through the previously established relationships. While still manual, the process can be assisted by software which performs this task through an intuitive interface which helps streamline the process, allows point-and-click measurements, identifies points or curves by their color, etc. An example of such software is WebPlotDigitizer \cite{webplotdigitizer} or PlotDigitizer \cite{plot_digitizer}. However, even with such software, the process is tedious and involves significant manual effort, with studies reporting an average time of up to fifteen minutes for data extraction from a single plot \cite{weblot_reliability}.

Therefore, automated and programmatic methods have been explored to overcome these limitations. Classical computer vision approaches rely on edge detection, Hough transforms, color-based segmentation, or object detection to identify axes and locate data points \cite{automated_program}, while more recent methods employ machine learning (ML), including neural networks \cite{neuralnetwork,cnn,chartocr}, robotic process automation \cite{milk} or transformer-based architectures, to segment lines and detect points in complex plots \cite{deplot}. Despite these advances, such solutions have not gained widespread adoption for several reasons. These techniques often rely on extensive parameter tuning, domain-specific knowledge, or labeled training data. Moreover, the interfaces and workflows for these methods can be cumbersome, requiring researchers to navigate complex software or perform multiple preprocessing and postprocessing steps. In practice, the difficulty and time investment associated with such methods mean researchers often opt for manual digitization procedures which are labor-intensive but are more straightforward in a day-to-day research workflow.
As a result, manual or semi-manual methods remain the norm for extracting data from plots in most research workflows.

Now, with large language models being multimodal, i.e. able of analyzing data other than pure text, such as images, new possibilities are opened in the field of extracting data from plots. In this paper we propose a zero-shot method of extracting data from research plots (digitization of plots) called \pe (\texttt{PE}). We consider only plots with two axes in this work, although generalization to more complex plots is likely straightforward. \texttt{PE} not only extracts the data in a convenient, standardized numerical format, but also verifies the success of the data extraction by comparing the extracted result to the source image. \texttt{PE} exploits multiple advanced LLM capabilities, including starting with leveraging the vision capabilities to analyze the plots, then python code generation to reproduce the plot followed by code output analysis to fix potential errors, and finally back to LLM vision again to evaluate the success of data extraction through visual comparison of two plots. We achieve an average relative error in the data extraction on published examples of around 5\% or less, suggesting that our approach is accurate and provides an effective tool for automated data extraction from images. Since the extraction is fast and relatively inexpensive \texttt{PE} can enable high-throughput automated data extraction from publications. The accuracy of the approach is expected to further improve as new and more capable multimodal LLMs are released.

\section{Description of the approach}
\label{sec:description}
\begin{figure*}
    \centering
    \includegraphics[width=0.9\textwidth]{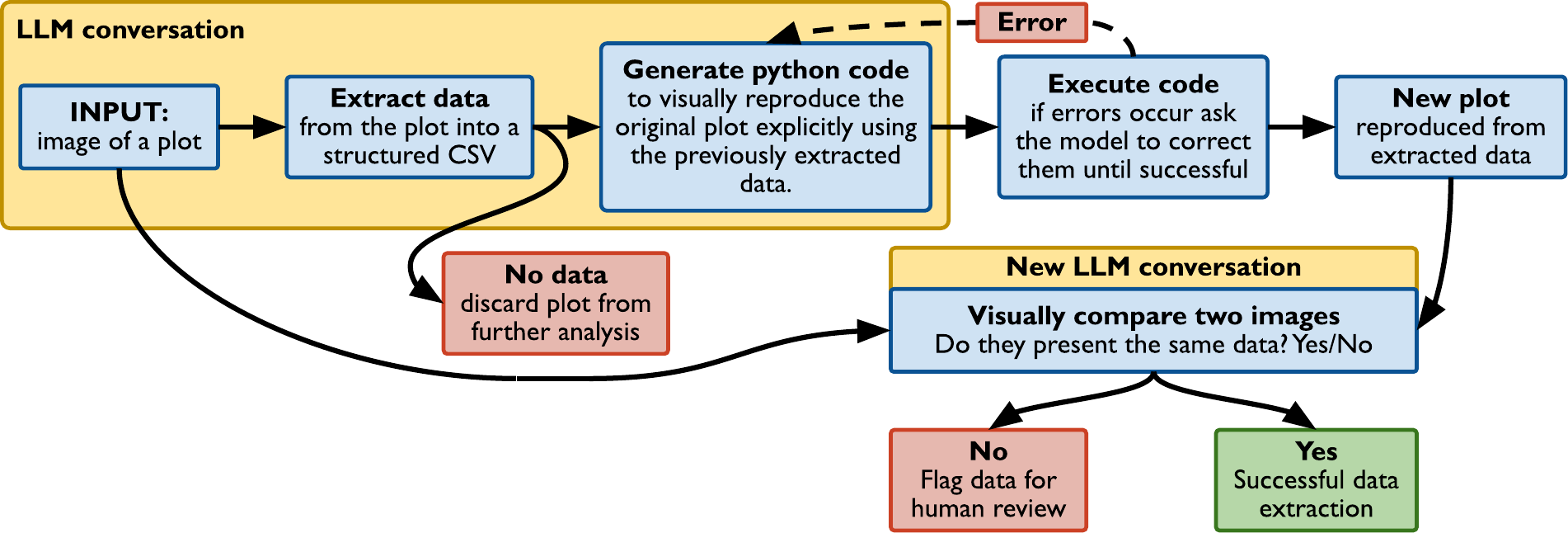}
    \caption{A simplified schematic of the vision LLM-based plot data extraction workflow.}
    \label{fig:simple}
\end{figure*}

Figure \ref{fig:simple} outlines the general workflow \texttt{PE} uses to extract data from plots. Figure \ref{fig:example} shows a visual example of the results fo the workflow. In this work we utilized Anthropic's Claude 3.5 Sonnet (see. Sec. \ref{sec:methods} for details) as the LLM, but in principle any vision-capable LLM could be used. Our exploration of the topic indicated that Claude 3.5 Sonnet performed better on the task than GPT-4o, so we decided to conduct the study and perform full evaluation for Claude 3.5 Sonnet. We describe the process in the following 4 steps:
\begin{enumerate}
\item \textbf{Extract data from the input plot image.} An image of the plot, alongside with a data extraction prompt is given to the LLM. The prompt precisely describes the data extraction task, as well as the required format of the extracted data, and directs the LLM to return the data only, with no other context. The prompt also explicitly allows the LLM to not return any data if there is none to be extracted. Note that this prompt is totally general and does not need to be altered for different types of images.
\item \textbf{Generate replotting code.} A prompt requesting the LLM to generate python code to reproduce the plot visually from the extracted data is given to the LLM. The prompt explicitly provides the previously extracted data back to the model to be used in recreation of the plot.
\item \textbf{Execute the generated code.} The code generated by the LLM is executed. If the code fails to execute, the entire error message is given back to the model along with an instruction to fix the previously generated code to account for the error, and the newly generated code is executed again. This process is repeated until the code executes, which occurred within 1-2 iterations in all cases we tested. This step generates what we call the "extracted plot".
\item \textbf{Compare original and extracted plot images.} A new conversation is started, in which the LLM is presented with the original plot image and the extracted plot image, alongside with a prompt asking the LLM to visually compare the two images and check whether they represent the same (or almost the same) data. A strict "yes" or "no" binary classification response is requested.
\end{enumerate}

The key aspect of automated data extraction is ensuring that it is accurate. Two steps in our workflow aim at mitigation of data extraction inaccuracies. In step 1 above, when prompting the model to extract the data, the model is explicitly allowed to respond that no data can be extracted. This may happen, for example, if the provided image is not a plot with extractable datapoints, key information such as an axis label is missing, or the plot is otherwise unreadable. The 4th step adds another accuracy check by ensuring that, if data is extracted, it is the same (or at least very similar) to the data in the source plot. That comparison includes all: axis ranges and labels, number of points/lines, and the overall position of points/lines and trends in their behavior. This step corrects mistakes in extraction (increasing overall precision) and catches cases of unreadable plots that managed to pass the screening in step 1. These steps are key in ensuring high accuracy, as each data point extracted from an unextractable plot would be, by definition, incorrect. The redundancy of the two step plot verification process maximizes precision of plot classification. This increases the potential for lower recall, by occasionally classifying an extractable plot as unextractable, but as seen in Tab. \ref{tab:statistics}, it does not happen often. Furthermore, we believe that in most applications precision is more important than recall. Precision errors yield incorrect data that is not directly flagged during the data extraction process, and could negatively impact later uses of the data. Recall errors associated with an incorrectly rejected image represent data that is never extracted and the original image is naturally flagged during the data extraction process as problematic in some way. Therefore, such errors do not pose a large risk of contaminating the data for later applications, and can be easily corrected by examining rejected figures for mistakes. In particular, when an image is incorrectly rejected it is almost always in step 4, at which point the data is already extracted, flagged by the model, and set aside for human review, and is likely correct and easily reclassified manually.

\section{Performance evaluation}

In this section we evaluate the accuracy of the data extraction. The method for such assessment is not obvious because of difficulties in comparing original and extracted data, and some choices must be made in the approach. For example, if a plot contains 3 datapoints (ground truth), and an extraction method extracts data on 2 points, how does one decide which 2 points from the ground truth to compare to? Shall these be the first two, the two that are closest to those extracted, or perhaps something else? Similar uncertainty in terms of the comparison procedure arises for more complex situations, e.g. in situations where more points than present in the ground truth are extracted. For the purpose of this paper we developed two consistent approaches to perform this type of comparison. This first we call \textit{pointwise comparison} and the second \textit{interpolation comparison}. We describe each in the text below. In all cases the ground truth are data points we carefully extracted manually from each image used in the assessment. Another important assessment is whether \texttt{PE} can determine if the source plot contains extractable data, i.e. whether the data on the plot is presented in a way that allows for precise data extraction (e.g. properly labeled curves, axis, tickmarks, etc.). We report this as \emph{plot classification} in \ref{tab:statistics}, and this performance is essential to the \pe approach, with two steps (step 1. and 4.) in the workflow (Sec. \ref{sec:description}) working together to maximize precision.

\subsection{Pointwise comparison}
\label{sec:pointwise}

\begin{figure*}
    \centering
    \includegraphics[width=0.9\textwidth]{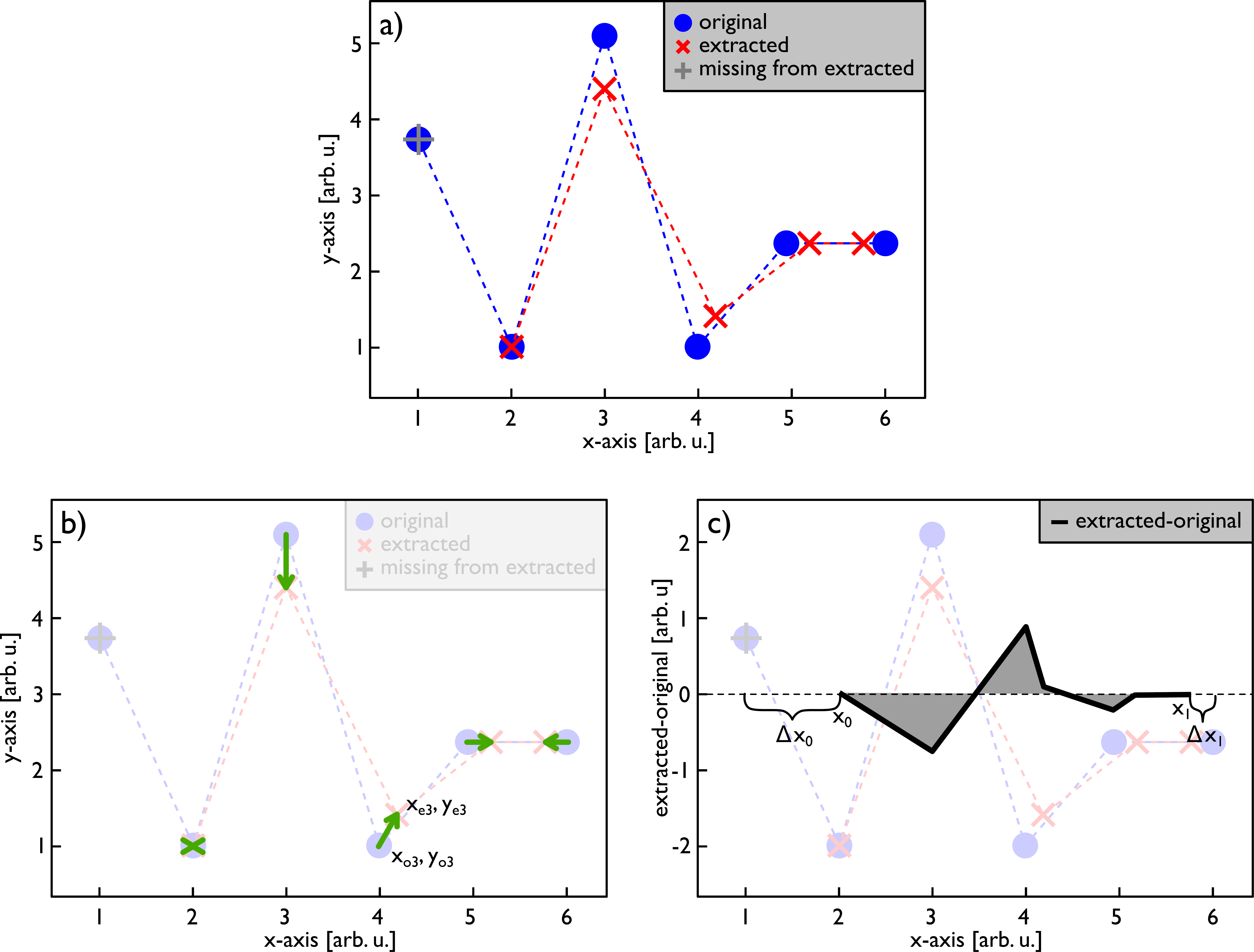}
    \caption{A schematic depiction of the two evaluation methods. Panel (a) shows the original and extracted datapoints on a single plot, connected with dashed lines to guide the eye. Panel (b) shows a pointwise comparison, and panel (c) shows an interpolation comparison.}
    \label{fig:compare}
\end{figure*}

This comparison began by calculating the ranges of \emph{x} and \emph{y} values for normalization. Then normalized Euclidean distance between any two points from the ground truth and extracted data were calculated, taking into account the combined ranges of both datasets to ensure consistent scaling. 
Then the pair of points (one from each dataset) with the smallest Euclidean distance in this normalized space was considered a match, and removed from further analysis. Iterations were performed until all points from the shorter of the datasets were matched, or until the leftover points corresponded to the beginning (lowest \emph{x} values) of one dataset and the end (highest \emph{x} values) of the other (which was a rare occurrence) in order to prevent matching obviously unrelated points. Then the mean absolute error (MAE) was calculated in both \emph{x} and \emph{y} directions between all paired points, normalizing these errors by the respective ranges of the data to provide a relative measure of discrepancy. The leftover unpaired points in both the original and extracted data were counted and reported as well in the precision and recall. Specifically, each point present in the ground truth and not matched with an extracted point was considered a false negative, and each point present in the extracted data without a match to ground truth was a false positive. A schematic of this comparison is given in Fig.~\ref{fig:compare} (b). The MAE has been calculated with $\text{MAE}_x = \frac{\sum |x_{\text{en}} - x_{\text{on}}|}{n}$ and $\text{MAE}_y = \frac{\sum |y_{\text{en}} - y_{\text{on}}|}{n}$, where the indexes $e$ and $o$ correspond to extracted and original datapoints, respectively. Precision and recall have been calculated as $\text{Precision} = \frac{n_e(\text{matched})}{n_e(\text{matched}) + n_e(\text{unmatched})}$ and $\text{Recall} = \frac{n_e(\text{matched})}{n_o}$.

\subsection{Interpolation comparison}
\label{sec:interpolation}

Another way of comparing the overall extracted data, more straightforward to describe and execute than pointwise comparison, is a comparison of lines created through interpolation of extracted datapoints and interpolated ground truth. We call this comparison \textit{interpolation comparison}. The motivation for this comparison is that it better describes how well the extracted data follows an overall trend of the curve, which in some cases may be the more desirable quantity. This comparison method also suffers from issues (e.g. it may incorrectly assess cases where just one extracted point is significantly off, with all others extracted perfectly, or curves close to straight lines match perfectly even when a significant number of points are missed by the extraction) but these are different than the issues in the \textit{pointwise comparison} and this approach therefore represents a complementary form of assessment.
The \textit{interpolation comparison} is rather straightforward. 
A script identifies the segments of the ground truth and extracted datasets that do not overlap in their independent variable \emph{x} range, quantifying the extent of the missing data on both the left and right ends. It calculates the left miss as the difference between the minimum \emph{x} value of the extracted dataset and the minimum \emph{x} value of the ground truth dataset, and similarly, the right miss as the difference between the maximum \emph{x} value of the ground truth dataset and the maximum \emph{x} value of the extracted dataset. To analyze the overlapping region, a finely spaced (1000 points) common uniform grid on the \emph{x} axis is generated within the bounds of the overlapping range to enable precise interpolation. Both datasets' dependent variable (\emph{y}) values are then interpolated onto this common \emph{x} axis using linear interpolation, providing a direct comparison between the datasets over the overlapping range. The absolute differences between the interpolated \emph{y} values at each point on the common axis are calculated. To quantify the average discrepancy in the overlapping region, the mean absolute error (MAE) is calculated as the mean of these absolute differences, formally speaking, its a numerical evaluation of $\text{MAE}_y = \int_{x_0}^{x_1} (\text{extracted}(x) - \text{original}(x)) \, dx / (x_1 - x_0)$. In addition, the error in $x$ is calculated based on the differences in where the plots start and ends: $\text{MAE}_{\Delta x} = (|\Delta x_0| + |\Delta x_1|) / 2$.

\subsection{Performance of the approach on test data}

We prepared two plot datasets for testing. The first consisted of plots generated by us from synthetic data, where we had full control over the appearance of the plot and the data itself, therefore we were able to introduce variability and accurate assess performance. The second data data set consisted of published plots selected randomly from research papers and manually digitized to provide ground truth data. More details about both plot datasets can be found in Sec. \ref{sec:datasets}.
On the synthetic plot dataset, the pointwise accuracy of the extracted data on a per point basis on the \emph{x} axis was a mean absolute error ($\text{MAE}_x$) of 2.9\%, and on the \emph{y} axis was a mean absolute error $\text{MAE}_y$ of 1\%. The precision and recall were both 89.3\%. Per plot results of pointwise accuracy were very similar with $\text{MAE}_x$ of 3.0\%, $\text{MAE}_y$ of 0.94\%, precision of 92.2\%, and recall of 91.9\%. 
Comparison of interpolated plots yielded a $\text{MAE}_y$ of 5.2\%, with the $\text{MAE}_{\Delta x}$ = 4.3\%.
The results of pointwise comparison per point on the published plot dataset proved to be accurate within a $\text{MAE}_y$ of 2.6\% and $\text{MAE}_x$ of 2.7\% with a precision and recall of 92.7\% and 91.3\% respectively. On a per plot basis results are comparable, with $\text{MAE}_x$ of 2.8\%, $\text{MAE}_y$ of 2.4\%, precision of 94.3\%, and recall of 92.6\%. The comparison on interpolated plots resulted in a $\text{MAE}_y$ of 3.4\% and $\text{MAE}_{\Delta x}$ 4.4\%.
In terms of identification of plots that contain extractable data, on both the published and synthetic datasets precision was 100\%, with recall of 81.8\% and 88.9\% respectively.

The synthetic data was much more challenging than published data due to its entirely random character, which often resulted in a sub-optimal presentation of the data on the plot, while published plots are designed to be clear and readable. This suggests that the statistics we obtained on the synthetic plot dataset represent a worst-case scenario for the accuracy of the approach.

\begin{table}[]
\begin{tabular}{lclcl}
\multicolumn{1}{r}{} & \multicolumn{2}{c}{\textbf{Synthetic Data}} & \multicolumn{2}{c}{\textbf{Published Data}} \\ \midrule
\multicolumn{5}{l}{\textbf{Plot classification}} \\ \midrule
Precision & \multicolumn{2}{c}{100\%} & \multicolumn{2}{c}{100\%} \\
Recall & \multicolumn{2}{c}{88.9\%} & \multicolumn{2}{c}{81.8\%} \\ \midrule
\multicolumn{5}{l}{\textbf{Interpolation comparison} (per plot)} \\ \midrule
MAE$_y$ & \multicolumn{2}{c}{5.2\%} & \multicolumn{2}{c}{3.4\%} \\
MAE$_{\Delta x}$ & \multicolumn{2}{c}{4.3\%} & \multicolumn{2}{c}{4.4\%} \\ \midrule
\multicolumn{5}{l}{\textbf{Pointwise comparison}} \\
\multicolumn{1}{c}{\textbf{}} & Per Plot & Per Point & Per Plot & Per Point \\ \midrule
Precision & 92.2\% & 89.3\% & 94.3\% & 91.7\% \\
Recall & 91.9\% & 89.3\% & 92.6\% & 91.3\% \\
MAE$_y$ & 0.94\% & 1.0\% & 2.4\% & 2.6\% \\
MAE$_x$ & 3.0\% & 2.9\% & 2.8\% & 2.7\%
\end{tabular}
\caption{Statistics for the accuracy of data extraction from plots. Classification relates to whether plots that are possible to unambiguously extract data from are classified correctly as such (i.e. whether plots impossible to extract data from are appropriately identified as such). Pointwise comparison is described in Sec.~\ref{sec:pointwise}. Interpolation comparison is described in Sec.~\ref{sec:interpolation}.}
\label{tab:statistics}
\end{table}

The major source of inaccuracies in data extraction seems to have come from the inherent nature of the LLM to "continue" the previous text. LLMs are very good at identifying and following patterns. Therefore, if data points of a certain plot have x-coordinates of "0, 1, 2, 3, 4.1, 5, 6", the model has a very hard time identifying the "4.1" value accurately, and very often will report it simply as 4, since this is the expected pattern to follow. The same also occurred for the \emph{y} coordinate. However, patterns occurred more rarely in the \emph{y} coordinate since the dependent value is usually more unevenly spaced.

\section{Conclusions}

In this study we proposed an approach for an effective leveraging of vision capabilities of multimodal large language models (LLMs) in automated extraction of data from research plots, a task that has traditionally been performed manually. By integrating the vision and language processing capabilities of LLMs with a zero-shot prompt engineering approach, this work demonstrates that accurate and efficient data extraction is achievable. We proposed a four-step workflow which combines visual data extraction, automated code generation and execution to reproduce the source plot using the extracted data, and a visual comparison of the source and reproduced images to ensure the accuracy of the extraction. This process reduces errors by explicitly allowing the model to reject unreadable plots while confirming the extracted data via replotting and comparison.

Performance evaluation on synthetic and published plot datasets revealed that the method achieves average errors of 4-6\% on synthetic plots and 3-5\% on published plots. At the same time, all plots with insufficient information for proper data extraction were identified and omitted, with only a limited amount of relevant plots being discarded, resulting in both high precision (100\%) and recall ($\approx$85\%). The randomization of plot styles, data structures, and labeling in synthetic plot datasets ensures that the approach is broadly applicable across diverse research domains. The synthetic plot dataset presented challenging scenarios designed to test the method’s limits, and the slightly better performance on published data demonstrates that our synthetic data results are likely a worst-case scenario.  

The study demonstrates the potential of multimodal LLMs to transform data extraction processes in research. As these models improve and new capabilities are introduced, we anticipate further reductions in error rates and enhanced generalization to more complex and specialized plot formats.

This work represents a significant step forward in automating data extraction from research plots, offering a robust, high-throughput alternative to manual methods. By enabling researchers to efficiently access quantitative data from plots in published works, this method enhances accessibility to scientific information and fosters new opportunities for data-driven insights across diverse fields.

\section{Plot Datasets}
\label{sec:datasets}
Published data extraction was assessed on a set of randomly selected 50 plots from research papers. ScienceDirect was used with a very broad search phrase \emph{"measurement"+"simulation"+"experimental"+"theoretical"}, from which the figures from the first 2000 results were downloaded, and 50 plots were randomly selected.
One unexpected challenge of the manual verification of the extracted data was that a surprising number of published plots do not allow for manual data extraction of a ground truth due to issues in the plot. For example, lines or points are often obscured by other lines or points, text, or legends. Sometimes axes are labeled in a way that makes it impossible to understand what the plot values represent, e.g. numbers not corresponding to any tickmarks, numbers appearing in a non-consecutive order, or missing entirely. 
Many of these plots still serve a useful purpose in the paper, e.g., a qualitative representation of a certain behavior, but these plots cannot be used for quantitative data extraction and assessment of \texttt{PE} accuracy. Out of the 50 plots, we were able to manually digitize only 44. However, the entire set of 50 plots was used for assessment to validate the behavior of the approach not only when presented with clear and correctly executed plots, but also ones impossible to analyze. Including both is important as it represents the situation that will likely be encountered when analyzing published data.

The synthetic plot dataset consists of 100 plots, 10 of which were purposefully edited to be impossible to accurately extract even manually (by removing axis, labeling axis with non-consecutive numbers, obscuring points behind other points so that they are indistinguishable, etc.). This dataset was designed to test various aspects of how plots may be presented in research, as well as to decouple the model's inherent knowledge from the data analysis i.e. assuming certain ranges of values typical for certain properties. This has been achieved by randomizing data, and corresponding labels, as well as the plots' appearance. In terms of the plots appearance the following properties have been explored through random sampling: plot type (points, lines, lines+points), colors of the points and lines, line thickness and point size, point shape and line style, font face, font size, presence of top and right plot axis, presence of grid, and plot size. In terms of the data, the following variations have been explored through random sampling: number of datasets presented on a single plot, scale of the data on both $x$ and $y$ axis, density of points used for the plot and whether the spacing is uniform or random. In addition axis labels (from a set of random names of properties) and units (from a set of random units) were randomly assigned as well, purposefully matching random units to random properties and to random curves, which was done in order to discourage model from trying to infer information from its inherent knowledge (e.g., expected order of magnitude of values for certain properties, or expected units for given properties). The plotted datapoints were variously scaled standard mathematical functions with noise added (again, to discourage the model from inferring the values based on the function rather than reading it from the plot). The used functions were sine, log, linear, gaussian. The use of mathematical functions was only to provide a rough trend to the points and lines, and the randomized axis values combined with added noise made the mathematical functions hard to distinguish - simulating an expected behavior of a variety of data present on plots in the literature. Full quantitative details of the data and plot appearance randomization process is given in the supplementary information (\cite{figshare}).

\section{Methods}
\label{sec:methods}
The Anthropic API was employed to generate conversational responses using the \texttt{claude-3-5-sonnet-20241022} model. The input parameters were configured with a temperature set to zero to produce more deterministic outputs. No explicit limit was set for max\_tokens, allowing the model to generate up to its inherent maximum. The interaction followed the message structure described in Fig. \ref{fig:simple} and explicitly detailed in the Supplementary Information. 
No system prompt was used. The API version was specified as 0.34.2 and the code was executed with python 3.9.6.

\section*{Acknowledgments} D.M and M.P.P. acknowledge the support from National Science Foundation Cyberinfrastructure for Sustained Scientific Innovation (CSSI) Award No. 1931298.

\section*{Data availability}
\label{sec:data}
All data and results of this research are publicly available and can be found on figshare \cite{figshare}. It includes the core \pe code (where details on the workflow and used prompts can be found), datasets, results, transcripts of LLM prompts and answers for each plot, ground truth used for comparison, and full accuracy analysis for all individual plots, as summarized in \ref{tab:statistics}. 

\section*{Author Contributions Statement} M. P. P. and D. M conceived the study. M. P. P. performed the modeling, tests and prepared/analyzed the results, D. M. guided and supervised the research. Writing of the manuscript was done by M. P. P. and D. M.. The authors acknowledge the help from Anna Latosinska in manually extracting ground truth data for model evaluation.
\section*{Competing Interest}
The authors declare no competing interests. 

\bibliography{biblio}
\end{document}